\documentclass{article}

\usepackage[dblblindworkshop, final]{neurips_2025}
\usepackage[utf8]{inputenc} % allow utf-8 input
\usepackage[T1]{fontenc}    % use 8-bit T1 fonts
\usepackage{hyperref}       % hyperlinks
\usepackage{url}            % simple URL typesetting
\usepackage{booktabs}       % professional-quality tables
\usepackage{amsfonts}       % blackboard math symbols
\usepackage{nicefrac}       % compact symbols for 1/2, etc.
\usepackage{microtype}      % microtypography
\usepackage{xcolor}         % colors

\usepackage{amsmath}
\usepackage{graphicx}
\usepackage{multirow}
\usepackage{colortbl}
\usepackage{xcolor}
\usepackage{color}
\usepackage{makecell}
\usepackage{marvosym}

\newcommand{\rc}[1]{{\scriptsize\color{magenta}{#1}}}

\title{InsFusion: Rethink Instance-level LiDAR-Camera Fusion for 3D Object Detection}
\workshoptitle{UrbanAI-2025: Harnessing Artificial Intelligence for Smart Cities}

\author{%
Zhongyu Xia$^{1}$\quad Hansong Yang$^{2}$\quad Yongtao Wang$^1$\textsuperscript{\Letter}
\\
$^1$Wangxuan Institute of Computer Technology, Peking University
\\
$^2$Beijing Jiaotong University
\\
\texttt{ \{xiazhongyu,wyt\}@pku.edu.cn \quad yanghansong@bjtu.edu.cn }
}

\begin{document}

\maketitle
\begin{abstract}
% LiDAR-camera fusion is critical for 3D object detection, yet existing methods face two flaws: (1) image-to-BEV (img2bev) relies on error-prone depth estimation, causing misaligned, low-quality image BEV features; (2) these errors propagate through early fusion and amplify in detection, degrading localization and classification. We propose InsFusion, a universal paradigm: by enabling each modality to optimize through independent interaction with ground truth (GT), it mitigates cumulative errors caused by depth estimation. Experiments on the nuScenes benchmark validate InsFusion: when integrated into FocalFormer and IS-Fusion, it delivers new state-of-the-art (SOTA) performance for LiDAR-camera fusion, underscoring its superiority and broad applicability.
Three-dimensional Object Detection from multi-view cameras and LiDAR is a crucial component for autonomous driving and smart transportation.
However, in the process of basic feature extraction, perspective transformation, and feature fusion, noise and error will gradually accumulate.
To address this issue, we propose InsFusion, which can extract proposals from both raw and fused features and utilizes these proposals to query the raw features, thereby mitigating the impact of accumulated errors.
Additionally, by incorporating attention mechanisms applied to the raw features, it thereby mitigates the impact of accumulated errors.
Experiments on the nuScenes dataset demonstrate that InsFusion is compatible with various advanced baseline methods and delivers new state-of-the-art performance for 3D object detection.
\end{abstract}    
\section{Introduction}
\label{sec:intro}

\begin{figure}[h]
    \centering
    \includegraphics[width=0.9\linewidth]{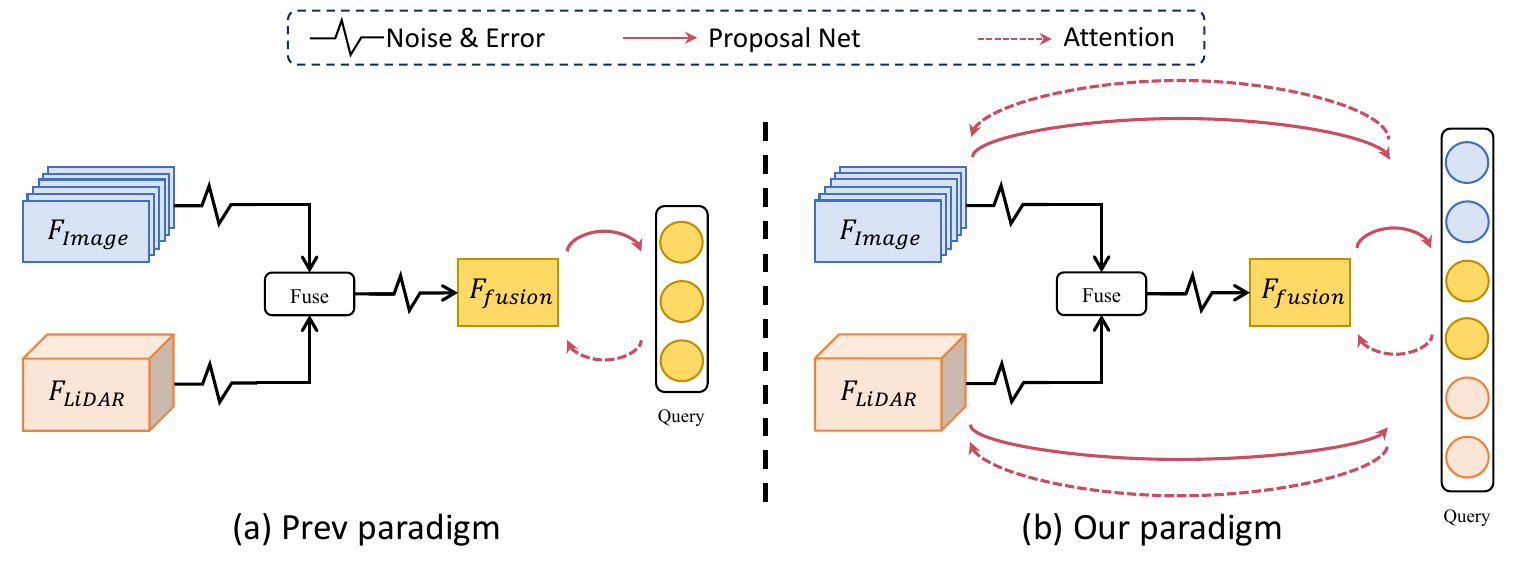}
    \vspace{-5pt}
    \caption{
    Comparison between the existing paradigm and our paradigm.
    %InsFusion can extract proposals from both raw and fused features, and utilizes these proposals to query the raw features, thereby alleviating the impact of accumulated errors.
%(a) Conventional Fusion Paradigm:
%The detection head queries features solely from the fused BEV representation and refines predictions based only on this error-accumulated feature stream. (b) Proposed InsFusion Paradigm:
%The detection head concurrently queries features from three distinct sources: the original LiDAR features, image features, and fusion features. It then jointly refines predictions using all three feature streams, effectively mitigating the influence of cumulative errors propagated through sequential processing.
}
    %\vspace{-10pt}
\end{figure}

% In the realm of autonomous driving, accurate 3D object detection is crucial. 
% The prevailing approach for high-performance 3D detection involves LiDAR-camera fusion, 
%
Multi-sensor fusion is essential for an accurate and reliable perception system, where the workflow typically entails extracting raw features from images and point clouds, transforming the 2D features into 3D representations, converting the 3D features from each modality into a consistent coordinate frame, fusing them, and extracting instance features from multimodal features. 
At each step of this process, various forms of noise or errors can be introduced. 
For instance, these may include depth estimation errors during 2D-to-3D transformation or offset prediction inaccuracies in deformable attention mechanisms, inaccuracies in extrinsic parameters during coordinate transformation, as well as information loss during the fusion process. These errors can accumulate and spread, ultimately affecting overall performance.
%typically entails generating Bird's-Eye View (BEV) representations from each modality separately and then fusing these BEV features. 
%
%Detection is subsequently performed on the fused BEV features. However, this widely adopted method is marred by two significant drawbacks (as illustrated in Fig 1.). Firstly, the img2BEV process hinges on depth estimation, a procedure prone to errors that can distort the image-derived BEV features. Secondly, these inaccuracies propagate and exacerbate during the early fusion stage, ultimately compromising the detection accuracy.

To address this issue, we introduce InsFusion, a universal instance-level fusion paradigm specifically tailored for Bird's-Eye View (BEV) LiDAR-camera fusion models. 
InsFusion extracts proposals from both raw and fused features and utilizes these proposals to query the raw features, thereby alleviating the impact of accumulated errors.
InsFusion can be easily integrated into all existing BEV-based LiDAR-camera fusion models, enhancing their performance.
%
% We conducted experiments on advanced methods, IS-Fusion~\cite{} and FocalFormer3D~\cite{}, on the nuScenes benchmark. 
%
% The results demonstrate that InsFusion significantly improves their performance, establishing a new state-of-the-art.

%Validated on the nuScenes benchmark with FocalFormer and IS-Fusion, InsFusion delivers consistent gains over baselines and achieves new state-of-the-art (SOTA) LiDAR-camera fusion results, which confirming its ability to mitigate error accumulation.

Our contributions are: 
(1) We propose InsFusion, a novel paradigm compatible with existing multimodal 3D object detection methods, to address the issue of accumulated errors.
(2) InsFusion requires only minimal fine-tuning of the baseline models, resulting in low training cost.
(3) InsFusion enhances the performance of both advanced models~\cite{chen2023focalformer3d, yin2024fusion} and achieves state-of-the-art results on the nuScenes dataset~\cite{Caesar2020nuScenesAM}.
 
%(1) We propose InsFusion, a to address cumulative errors; (2) Showing its universal applicability, which boosts FocalFormer and IS-Fusion to SOTA on nuScenes; (3) Validating its low training cost, as it only requires fine-tuning without full retraining.

\section{Related Work}

\textbf{LiDAR-only.}
Prior LiDAR-based works directly operate on the raw LiDAR point clouds~\cite{qi2017pointnet++,Qi2017PointNetDL,qi2018frustumPF,Shi2019PointRCNN3O,Yang20203DSSDP3,li2021lidar} to extract features.
With the growing volume of LiDAR points in outdoor datasets, several works transform point clouds into Euclidean feature space, such as 3D voxels~\cite{Zhou2018VoxelNetEL}, range images~\cite{fan2021rangedet, Sun2021RSNRS}, and bird's eye view (BEV) plane~\cite{Lang2019PointPillarsFE, Wang2020PillarbasedOD, Yin2020Centerbased3O, fan2021rangedet, Sun2021RSNRS}.

\textbf{Camera-only.}
Early works~\cite{lu2021geometry, Roddick2019OrthographicFT, liu2021autoshape, kumar2021groomed, zhang2021objects, zhou2021monocular, reading2021categorical, wang2021progressive, wang2021depth} predicted 3D bounding boxes from monocular images. 
With the introduction of datasets such as nuScenes~\cite{Caesar2020nuScenesAM}, methods for multi-view cameras have demonstrated superior performance.
The core problem they address involves transforming 2D feature maps into 3D features while extracting instance features.
LSS~\cite{philion2020lift} and related methods~\cite{huang2021bevdet, li2023bevdepth, sts, HoP, SOLOFusion, xia2024henet} employ depth estimation to lift 2D features into the BEV space, subsequently extracting instance features through techniques such as heatmap peak detection~\cite{Yin2020Centerbased3O}.
BEVFormer series~\cite{bevformer, bevformerv2} acquires BEV features through Deformable Attention~\cite{zhu2020deformable}.
Subsequent approaches~\cite{wang2022detr3d, liu2022petr,liu2023petrv2, sparse4d, sparse4dv2, wang2023exploring, liu2023sparsebev} directly obtain instance features by querying 2D feature maps.
They streamline the feature extraction pipeline, thereby mitigating cumulative errors and consequently achieving superior performance in object detection tasks.

%Camera-based 3D detection faces depth ambiguity but has gained research interest for its cost-effectiveness, with methods split into two core paradigms: dense BEV construction and sparse query-based.
%
%Dense BEV methods transform image features to BEV via depth estimation or attention mechanisms. For example, LSS \cite{philion2020lift} uses depth distributions to project frustum features; BEVDet \cite{huang2021bevdet} improves feature resolution and inference speed with multi-scale BEV encoders; BEVDepth \cite{li2023bevdepth} incorporates LiDAR-generated depth supervision to boost depth accuracy;

% BEVFormer \cite{li2024bevformer} employs spatiotemporal cross-attention to model dynamic scenes.
% %
% Query-based methods avoid the computational overhead of dense BEV construction. DETR3D \cite{wang2022detr3d} projects 3D reference points onto multi-view images for feature sampling; In the PETR series \cite{liu2022petr,liu2023petrv2,wang2023exploring}, global attention is employed to facilitate the interaction between queries and features. Additionally, 3D positional embeddings are utilized to transform 2D features into 3D representations, bypassing the need for explicit projection; SparseBEV \cite{liu2023sparsebev} introduces hybrid query mechanisms and adaptive spatiotemporal aggregation to enhance long-range perception while maintaining sparse efficiency.

\textbf{LiDAR-camera fusion.}
In contrast to early-fusion~\cite{Huang2020EPNetEP, Sindagi2019MVXNet, Wang2021PointAugmentingCA, Yin2021MVP} or late-fusion~\cite{pang2020clocs} strategies, feature-level fusion methods have demonstrated superior performance.
Existing feature-level fusion methods~\cite{Yoo20203DCVFGJ, bai2022transfusion, liang2022bevfusion, liu2022bevfusion} are primarily extensions of LiDAR-only approaches, with BEV-based methods constituting the predominant paradigm.
Such paradigms typically cascade multiple modules, thereby introducing cumulative errors. Subsequent methods often aim to mitigate inaccuracies in specific components of this pipeline, such as reducing depth estimation errors~\cite{hu2023ea, zhao2024simplebev} to alleviate noise in the transformation from 2D feature maps to BEV features.
IS-Fusion \cite{yin2024fusion} employs instance-guided Fusion to mitigate noise introduction during the BEV fusion process.
FocalFormer3D \cite{chen2023focalformer3d} uses multistage heatmaps to reduce errors in the instance feature extraction process from multimodal BEV features.
In contrast to previous efforts that primarily focus on refining individual components within the pipeline, InsFusion leverages minimally processed raw features to mitigate noise while maintaining full compatibility with existing methods, thereby enhancing their performance.

% Multimodal fusion leverages the geometric precision of LiDAR point clouds and semantic richness of images, becoming a critical research direction with three main paradigms. Early fusion combines raw sensor data directly, such as projecting image semantics onto point clouds, but suffers from calibration errors and noise-induced feature distortion. Late fusion integrates detection results after independent modality processing, yet fails to exploit cross-modal complementarity because of absent feature interaction. Feature-level fusion is the dominant type, including methods like BEVFusion \cite{liang2022bevfusion,liu2022bevfusion} (which projects features into unified BEV space via dual-branch encoders), TransFusion \cite{bai2022transfusion} (which refines sparse LiDAR instance features with dense image cues), FocalFormer3D \cite{chen2023focalformer3d} (which mitigates false negatives through multi-stage heatmap prediction) and IS-Fusion \cite{yin2024fusion} (which advances state-of-the-art via instance-scene interaction).
%
%However, these methods inherit limitations from the legacy paradigm: inaccurate depth estimation during image-to-BEV projection degrades feature quality, while uncorrected error propagation from fusion directly compromises detection heads. In contrast, our novel approach fundamentally resolves these issues.
\section{Methodology}

% InsFusion extracts proposals from both raw modality features and fused features, then uses these proposals to query raw features and fused features. 
% %
% InsFusion consists of three parallel branches (Camera, LiDAR, Fusion) for proposal extraction, a modal projection module for unified feature space alignment, and a cross-modal decoder for integrated refinement.

\begin{figure}
    \centering
    \includegraphics[width=1.0\linewidth]{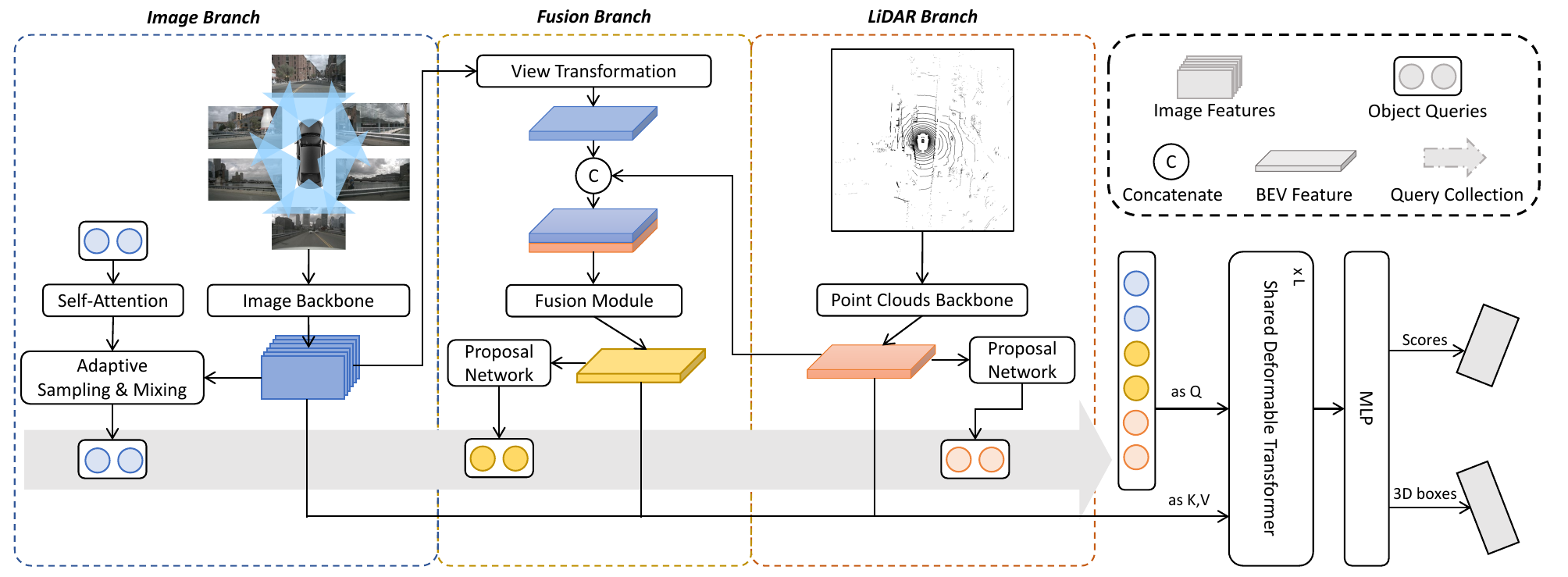}
    \caption{
    \textbf{Overview of the InsFusion framework.}
    The framework extracts proposals from raw camera features, LiDAR features, as well as fused BEV features, and then aligns and refines all proposals to predict 3D bounding boxes.
    %The framework extracts proposals from raw camera/LiDAR features and fused BEV features, then aligns and refines all proposals to predict 3D bounding boxes. %generate final 3D detection results.
}
\label{fig:method}
%\vspace{-20pt}
\end{figure}

As shown in Fig.~\ref{fig:method}, in order to extract information from features with minor cumulative noise, InsFusion takes three steps: extracting proposals from 2D feature maps, LiDAR features, and fused features; performing query alignment; and refining instance features through attention mechanisms.

\subsection{Extract Queries From Raw Features}

% Each branch is tailored to extract proposals from its source feature (camera 2D features, LiDAR features, or fused BEV features) and query raw features to ensure direct access to low-noise information, mitigating error accumulation from intermediate processing (e.g., dense BEV projection). The specific implementations are as follows:

\noindent\textbf{Camera Branch.} 
%
%Firstly, we use image backbone to obtain image features $F_{img}\in \mathbb{R}^{h\times w\times C_{img}}$ .To extract instance-level proposals from $F_{img}$and bridge 2D-3D alignment, we adopt the approach of adaptive sampling and mixing to obtain proposals of image features while avoiding the construction of dense BEV:
%
For image features $F_{img}\in \mathbb{R}^{h\times w\times C_{img}}$, 
we first define $K$ learnable camera queries, which are drawn from random Gaussian distributions and attached to a $D_q$-dim query feature to encode the rich instance characteristics.
Then we adopt adaptive sampling and mixing~\cite{liu2023sparsebev} to obtain instance features $Q_{img}^{(0)}\in \mathbb{R}^{K\times D_q}$.

%, where $K$ denotes the number of these learnable embeddings (i.e., the total number of base camera queries) and $D_q$ represents the feature dimension of each individual embedding. Thus, $Q_{img}^{(0)}$ is defined as $(Q_{img}^{(0)}\in \mathbb{R}^{K\times D_q}$, where each embedding explicitly encodes 3D instance priors: $[x, y, z, w, l, h, v_x, v_y]$ (representing 3D center coordinates, dimensions, and in-plane velocity). Then, instead of projecting the entire image into a dense BEV, we only sample image feature regions relevant to each $Q_{\text{img}}^{(0)}$  (guided by their 3D prior coordinates). This sparse interaction retains fine-grained semantic details of the image while avoiding error accumulation in 2D-to-3D projection, ultimately producing instance-level camera queries $Q_{\text{img}}$ for subsequent detection.

\noindent\textbf{Lidar Branch.} 
% Following TransFusion-L, we first process raw point clouds to obtain LiDAR BEV features $F_{lidar\_bev}\in R^{X\times Y\times C_{lidar\_bev}}$.
For LiDAR BEV features $F_{lidar\_bev}\in R^{X\times Y\times C_{lidar\_bev}}$, InsFusion predicts the instance heatmap and extracts proposals through peak detection.
%
%To initialize the LiDAR query $Q_{\text{lidar}}$, we follow the heatmap peak detection strategy in TransFusion-L: a heatmap prediction head outputs a LiDAR BEV heatmap $H_{\text{lidar}} \in \mathbb{R}^{X \times Y}$, where peak values indicate potential object centers. 
%
We select the highest K peaks from $H_{\text{lidar}}$ and encode their spatial coordinates into initial embeddings, forming the initial LiDAR query $Q_{\text{lidar}}^{(0)} \in \mathbb{R}^{K \times D_q}$. 
%
% It is refined through the layers of the transformer decoder $L_l$, with $F_{\text{lidar\_bev}}$ serving as Key-Value to aggregate geometric features (e.g., 3D position and dimension) from the LiDAR BEV space:

% \begin{equation}
%     Q_{\text{lidar}}^{(l)} = \text{DecoderLayer}( Q_{\text{lidar}}^{(l-1)}, Flatten(F_{\text{lidar\_bev}})),
% \end{equation}

% where $Flatten(.)$ denotes flattening operation, and the final instance-level LiDAR query is: $Q_{\text{lidar}} = Q_{\text{lidar}}^{(L)}$

\noindent\textbf{Fusion branch.} 
InsFusion is a framework compatible with existing multimodal fusion methods, applicable to any feature fusion strategy, whether it employs BEV heatmaps or Transformer-based architectures to obtain the fused instance features $Q_{\text{fusion}}^{(0)}$.

% First, we project the image feature $F_{\text{img}}$ into the BEV space using the Lift-Splat-Shoot (LSS) method, which estimates depth distributions for each image pixel and splats them into a BEV grid. This results in the camera BEV feature $F_{\text{img\_bev}} = \text{LSS}(F_{\text{img}}), \quad F_{\text{img\_bev}} \in \mathbb{R}^{X \times Y \times C_{\text{img\_bev}}}$ where $C_{\text{img\_bev}} = 256$, and $X, Y$ are consistent with the LiDAR BEV spatial dimensions to ensure alignment.
%
% Next, following a strategy analogous to BEVFusion, we fuse $F_{\text{img\_bev}}$ with $F_{\text{lidar\_bev}}$ via channel concatenation followed by a convolution layer to aggregate cross-modal information, yielding the fused BEV feature:
%
% \begin{equation}
%     F_{\text{fusion\_bev}} = \text{Conv}\left( \text{Concat}(F_{\text{lidar\_bev}}, F_{\text{img\_bev}}) \right).
% \end{equation}
%
% $\text{Concat}(\cdot)$ concatenates the features along the channel dimension.
%
% In practice, we initialize the fusion query $Q_{\text{fusion}}^{(0)}$ using the initial LiDAR query $Q_{\text{lidar}}^{(0)}$ (leveraging LiDAR's reliable geometric prior), Similar to the LiDAR branch, we employ the same Transformer decoder as used in the LiDAR branch for decoding, resulting in the final fusion query.${Q_\text{fusion}} = Q_{\text{fusion}}^{(L)}$

\noindent\textbf{Queries Alignment.}
%
%\subsection{Modal Projection for Unified Query Space}
In this stage, the three sets of queries ($Q_{\text{img}} \in \mathbb{R}^{K \times D_q}$, $Q_{\text{lidar}} \in \mathbb{R}^{K \times D_q}$, $Q_{\text{fusion}} \in \mathbb{R}^{K \times D_q}$) are projected into a shared latent space via modality-specific linear transformers, eliminating distribution shifts while preserving modality-specific strengths:
$\hat{Q}_{\text{img}} = W_{\text{img}} \cdot Q_{\text{img}} + b_{\text{img}}$, $\hat{Q}_{\text{lidar}} = W_{\text{lidar}} \cdot Q_{\text{lidar}} + b_{\text{lidar}}$, $\hat{Q}_{\text{fusion}} = W_{\text{fusion}} \cdot Q_{\text{fusion}} + b_{\text{fusion}}$,
%
% \begin{equation}
% \begin{aligned}
%     &\hat{Q}_{\text{img}} = W_{\text{img}} \cdot Q_{\text{img}} + b_{\text{img}}, \\
%     &\hat{Q}_{\text{lidar}} = W_{\text{lidar}} \cdot Q_{\text{lidar}} + b_{\text{lidar}}, \\
%     & \hat{Q}_{\text{fusion}} = W_{\text{fusion}} \cdot Q_{\text{fusion}} + b_{\text{fusion}},
% \end{aligned}
% \end{equation}
%
where $W_{\text{img}}, W_{\text{lidar}}, W_{\text{fusion}} \in \mathbb{R}^{D_q \times D_q}$ are learnable projection matrices, and $b_{\text{img}}, b_{\text{lidar}}, b_{\text{fusion}} \in \mathbb{R}^{1 \times D_q}$ are bias terms. These parameters are optimized end-to-end with the entire network.

\subsection{Instance Feature Refinement}

After alignment, we concatenate the three aligned queries sets (\(\hat{Q}_{\text{img}}, \hat{Q}_{\text{lidar}}, \hat{Q}_{\text{fusion}}\)) to form a comprehensive query \(Q^{(0)}\).
This design integrates low-error information from raw-modality proposals and cross-modal context from fusion proposals, enabling multi-source feature interaction during refinement.
Instead of using a single feature map as Key (K) and Value (V), we leverage the three core feature sources (raw camera features \(F_{\text{img}}\), raw LiDAR BEV features \(F_{lidar\_bev}\), and fused BEV features \(F_{fusion_bev}\)) as separate Key-Value. 
The concatenated query \(Q_{\text{concat}}\) performs targeted queries on each of these three feature sources via a deformable transformer decoder—this ensures the query pool fully captures modality-specific strengths (e.g., image semantics, LiDAR geometry, cross-modal complementarity) while avoiding error accumulation from over-reliance on a single feature source. The refinement process follows:

\begin{equation}
\begin{aligned}
&Q^{(0)} = \left[ \hat{Q}{\text{img}}, \hat{Q}{\text{lidar}}, \hat{Q}{\text{fusion}} \right] \in \mathbb{R}^{3K \times D_q}, \\
&Q^{(l)} = \text{DeformableTransformerLayer}( Q^{(l-1)}, \text{Flatten}(F_{\text{img, raw}}), \text{Flatten}(F_{\text{lidar, raw}}), \text{Flatten}(F_{\text{fusion}}) ).
\end{aligned}
\end{equation}

This process is repeated for $L$ iterations.
Here, \(\left[\cdot,\cdot,\cdot\right]\) denotes concatenation, and the deformable decoder layer independently computes attention between \(Q_{\text{concat}}^{(l-1)}\) and each of the three flattened Key-Value pairs (i.e., $F_{img}$, \(F_{lidar_bev}\), \(F_{fusion\_bev}\)). The attention outputs from the three feature sources are aggregated via element-wise addition to update the query.
This multi-source query design ensures InsFusion fully leverages the precision of raw features and the complementarity of fused features, directly addressing error accumulation by anchoring refinement to low-error, modality-specific sources.
\section{Experiments}

\begin{table*}[t]
\centering
% \vspace{-0.6cm}
\small
\setlength{\tabcolsep}{2pt}
\caption{
  \textbf{Results on the nuScenes {\tt val} set. }
  Applying InsFusion to FocalFormer and the state-of-the-art method IS-Fusion effectively enhances their performance.
  % When integrated with FocalFormer3D, InsFusion yields performance gains of +1.0\% mAP and +1.1\% NDS; when applied to the state-of-the-art method IS-Fusion, InsFusion achieves improvements of +1.1\% mAP and +0.6\% NDS.
}
\label{tab:nuscene_test}
\resizebox{\textwidth}{!}{
\begin{tabular}{l|cc|cc|cccccccccc}
    \toprule
     {Method} & \small{Mod.} & \small{I.B.} & \small{mAP} & \small{NDS} & \scriptsize{Car} & \scriptsize{Truck} & \scriptsize{C.V.} & \scriptsize{Bus} & \scriptsize{Trailer} & \scriptsize{Barrier} & \scriptsize{Motor.} & \scriptsize{Bike} & \scriptsize{Ped.} & \scriptsize{T.C.} \\
     \midrule
     BEVFormer~\cite{bevformer} & C & V2-99  & 41.6 & 51.7 & 61.8& 37.0& 12.8& 44.4& 17.2& 52.5& 42.9& 39.8& 49.4& 58.4\\
     BEVDet4D~\cite{huang2021bevdet} & C & R50 & 40.8 & 52.3 & 62.7& 33.4& 12.4& 39.3& 15.6& 58.7& 38.5& 39.1& 46.7& 61.8\\
     SparseBEV~\cite{liu2023sparsebev} & C & V2-99 & 57.3 & 65.0 &72.5 & 53.8& 25.9& 61.3& 33.3& 67.7& 58.9& 60.8& 65.2& 73.9 \\
     
     %\midrule
    
     CenterPoint~\cite{Yin2020Centerbased3O} & L & -  & 58.6 & 65.8 & 84.0 & 53.8 & 19.9 & 68.1 & 38.7 & 67.8 & 62.2 & 43.4 & 83.1 & 65.3  \\
     Transfusion-L~\cite{bai2022transfusion} & L & -  & 65.1 & 70.1 & 86.6 & 41.1 & 26.1 & 70.6 & 38.1 & 68.5 & 69.4 &54.7 & 85.5 & 71.9 \\
     FocalFormer3D-L~\cite{chen2023focalformer3d} & L &  - & 66.3 & 70.1 & 87.4 & 59.1 & 27.8 & 76.4 & 44.3 & 71.1 & 75.3 & 60.4 & 87.2 & 74.0 \\

    %\midrule
    {FUTR3D}~\cite{chen2022futr3d} &L+C&R101 &64.2 &68.0  &86.3  &61.5  &26.0  &71.9  &42.1  & 64.4 & 73.6 & 63.3 &82.6 &70.1 \\
    BEVFusion~\cite{liang2022bevfusion} & L+C & Swin-T &  69.6 & 72.1 & 89.1 & 66.7 & 30.9 & 77.7 & 42.6 & 73.5 & 79.0 & 67.5 & 89.4 & 79.3 \\

    \midrule
    FocalFormer3D~\cite{chen2023focalformer3d} & L+C &  R50 & 70.5 & 73.1 & \textbf{89.4} & 65.4 & 30.5 & 78.8 & 45.6 & 73.2 & \textbf{83.0} & \textbf{70.9} & 88.7 & 79.5 \\
    \textbf{+InsFusion}~(ours) & L+C & R50  & 71.5\rc{(+1.0)} & 74.2\rc{(+1.1)} & 88.9& \textbf{69.7}& 30.5& 77.7& \textbf{53.1}& \textbf{84.2}& 78.4& 65.0& 89.0& 78.9\\

    \midrule
    % IS-Fusion~\cite{yin2024fusion}\dag & L+C & Swin-T  & 72.8 & 74.0 & \\
    IS-Fusion~\cite{yin2024fusion}\ddag & L+C & Swin-T & 72.3 & 73.7 & \textbf{89.4} & 68.3 & 36.1 & 79.4 & 51.6 & 74.7 & 80.6 & 70.1 & 89.3 & 83.6\\
    \textbf{+InsFusion}~(ours)  & L+C & Swin-T & \textbf{73.4}\rc{(+1.1)}& \textbf{74.3}\rc{(+0.6)}& \textbf{89.4}& 68.1& \textbf{37.0}& \textbf{79.5}& 52.9& 80.1& 80.3& 70.4& \textbf{91.3}& \textbf{85.4}\\
    \bottomrule
    \multicolumn{14}{l}{\scriptsize{'Mod.': Modality, 'I.B.': Image Backbone. 
    %\dag~: Reported (official paper), 
    \ddag~: Reproduced with official code. Notion of modality: Camera (C), LiDAR (L).}}\\
    \multicolumn{14}{l}{\scriptsize{Notion of 
    class: Construction vehicle (C.V.), pedestrian (Ped.), traffic cone (T.C.).  }}\\
\end{tabular}
}
%\vspace{-3mm}
\end{table*}

We provide experimental setup and implementation details in the Appendix ~\ref{appendix:exp_setup}.

\textbf{Main Result.} 
Experimental results on the nuScenes~\cite{Caesar2020nuScenesAM} dataset are summarized in Tab. \ref{tab:nuscene_test}.
%, which demonstrate consistent improvements across two baseline methods. 
%
When integrated with FocalFormer3D, InsFusion yields performance gains of +1.0\% mAP and +1.1\% NDS; when applied to the state-of-the-art method IS-Fusion, InsFusion achieves improvements of +1.1\% mAP and +0.6\% NDS. These results show the effectiveness of InsFusion.
%
% FocalFormer3D+InsFusion achieves 74.2\% NDS (+1.1\%) and 71.5\% mAP (+1.0\%), while IS-Fusion+InsFusion achieves 74.3\% (+0.6\%) NDS and 73.4\% mAP(+1.1\%). These improvements provide conclusive evidence that our InsFusion model is effective.
%
Unfortunately, due to a service outage in the official nuScenes evaluation server, we were unable to evaluate the results on the test set.

\textbf{Ablation study for the number of Transformer layers.} 
%
% To verify how the number of layers (L) in the shared deformable transformer affects InsFusion's performance, we test \(L=1,2,6\), (all other hyperparameters, e.g.,\(K=300\), kept consistent).
%
As shown in Tab. \ref{tab:ablation_transformer_layers}, when \( L = 1 \), the single-layer deformable Transformer exhibits insufficient capability to refine concatenated proposals, which prevents the full capture of complementary information across multi-source features (raw camera, raw LiDAR, and fused BEV); and when \( L = 6 \), excessive iterative refinement induces the model to overfit to local noise generated during the fusion process rather than focusing on critical global instance-level information, thereby leading to performance degradation, \( L = 2 \) is ultimately identified as the optimal configuration, as it achieves a balanced trade-off between ensuring adequate feature refinement and avoiding overfitting to local noise.

\begin{table}[t]
    \centering
    \caption{Ablation study on the number of deformable transformer layers. The metrics mAP (\%) and NDS (\%) are computed on the nuScenes {\tt val} set.}
    \label{tab:ablation_transformer_layers}
    \begin{tabular}{ccc}
        \hline
        \textbf{Layers Number (\( L \))} & \textbf{mAP (\%)} & \textbf{NDS (\%)}  \\
        \hline
        1 & 67.31 & 71.35 \\
        2 & \textbf{69.33} & \textbf{72.35} \\
        6 & 68.82 & 71.59 \\
        \hline
    \end{tabular}
\end{table}

\section{Conclusion}
% 3D object detection with multi-view cameras and LiDAR is critical for autonomous driving and smart transportation, yet it suffers from noise and error accumulation during feature processing and fusion—degrading performance. 
% To solve this, we proposed InsFusion, a universal instance-level fusion paradigm for LiDAR-camera fusion. It extracts queriess from raw and fused features, queries raw and fused features to alleviate accumulated errors.
% InsFusion is compatible with existing BEV-based fusion models and only requires minimal fine-tuning (with low training cost). 
% Experiments on the nuScenes dataset demonstrate that InsFusion can further enhance the performance of several  high-performing baseline methods.
%NuScenes experiments confirm it enhances advanced baselines (e.g., FocalFormer3D, IS-Fusion) to new state-of-the-art 3D detection performance, offering a practical solution to advance multimodal 3D perception.

3D object detection using multi-view cameras and LiDAR is crucial for autonomous driving and smart transportation. 
However, it suffers from noise and error accumulation during feature processing and fusion, which degrades performance. 
To solve this, we proposed InsFusion, a universal instance-level fusion paradigm for LiDAR-camera fusion. It extracts queries from both raw and fused features, and then queries these features to alleviate accumulated errors.
InsFusion is compatible with existing BEV-based fusion models and only requires minimal fine-tuning (with low training cost). 
Experiments on the nuScenes dataset demonstrate that InsFusion can further enhance the performance of several high-performing baseline methods.

\section*{Acknowledgements}

This work was supported by the National Natural Science Foundation of China (Grant No. 62176007).

\bibliographystyle{unsrt}
\bibliography{ref}  

\clearpage
\appendix
\section{Experimental Setup.}
\label{appendix:exp_setup}
\textbf{Dataset}. Our experimental evaluation is performed on the nuScenes dataset, a standard benchmark for autonomous driving research that provides comprehensive multi-sensor data across diverse urban environments. This dataset enables rigorous evaluation of 3D detection systems under challenging real-world conditions.We adopt the official evaluation metrics: NuScenes Detection Score (NDS) and mean Average Precision (mAP), which comprehensively assess detection accuracy, localization precision, and orientation prediction.

\textbf{Implementation Details}. Our main implementation relies on the open-source MMdetection3D~\cite{mmdet3d2020} framework. When deploying InsFusion on the FocalFormer and IS-Fusion baselines, we maintain all the original settings of the baseline model, including the LiDAR backbone, camera backbone, and the original fusion network, to ensure a fair comparison.

For the key parameters of InsFusion: the number of layers for the Shared Deformable Transformer is set to 
$L=2$; the initial query count for both the LiDAR branch and camera branch is 300. We train our network using a two-stage training mode: Firstly, train the baseline model (skip this step if official weights are available and use the official weights directly), and then train the "extract proposals from img" network.
When training the InsFusion-enhanced model, load and freeze the baseline weights, and load the weights of the "extract proposals from img" network.

For optimization, we use the Adam optimizer with a one-cycle learning rate policy (max learning rate $2\times 10^{-5}$) and a weight decay of 0.01. All experiments are conducted on 8 NVIDIA RTX 8000 GPUs, with a batch size of 16 and a training duration of 6 epochs.

\section{Limitation.}
\label{appendix:limitation}
A limitation of InsFusion is that it introduces additional computational overhead compared to baseline models, which manifests as a slight reduction in inference efficiency. However, as shown in Tab. \ref{tab:computational_overhead}, this efficiency cost is insignificant and well within acceptable limits for practical use: when integrated into the FocalFormer baseline, the inference FPS decreases by only 9.3\%; when applied to the IS-Fusion baseline, the FPS reduction is even smaller, at just 6.6\%. Based on the inference FPS metric alone, the resulting decrease in inference latency remains within an acceptable range, indicating that its computational overhead does not impede practical deployment in 3D detection applications.

\begin{table}[h]
    \centering
    \caption{Computational Overhead Comparison Between InsFusion-Enhanced Models and Baselines}
    \label{tab:computational_overhead}
    \begin{tabular}{lcc}
        \hline
        \textbf{Model Name}            & \textbf{Inference FPS$\uparrow$}                 \\
        \hline
        FocalFormer            & 1.29         \\
        \textbf{+InsFusion}    & 1.17   \\
        \hline
        IS-Fusion              & 0.91    \\
        \textbf{+InsFusion}     & 0.85     \\
        \bottomrule
    \end{tabular}
    \vspace{1mm}
    
\end{table}

\section{Broader Impacts Statement.}
\label{appendix:braoder_impacts}
All datasets we used are published datasets. We do not see potential privacy-related issues. This study may inspire future research on LiDAR-camera fusion model for 3D perception.

\clearpage
\section*{NeurIPS Paper Checklist}
\begin{enumerate}

\item {\bf Claims}
    \item[] Question: Do the main claims made in the abstract and introduction accurately reflect the paper's contributions and scope?
    \item[] Answer: \answerYes{} % Replace by \answerYes{}, \answerNo{}, or \answerNA{}.
    \item[] Justification: The abstract and introduction clearly state the paper’s core contributions.

\item {\bf Limitations}
    \item[] Question: Does the paper discuss the limitations of the work performed by the authors?
    \item[] Answer: \answerYes{} % Replace by \answerYes{}, \answerNo{}, or \answerNA{}.
    \item[] Justification: We discuss the limitations of this work in Appendix. \ref{appendix:limitation}.

\item {\bf Theory assumptions and proofs}
    \item[] Question: For each theoretical result, does the paper provide the full set of assumptions and a complete (and correct) proof?
    \item[] Answer: \answerNA{} % Replace by \answerYes{}, \answerNo{}, or \answerNA{}.
    \item[] Justification: This paper does not include theoretical results.

    \item {\bf Experimental result reproducibility}
    \item[] Question: Does the paper fully disclose all the information needed to reproduce the main experimental results of the paper to the extent that it affects the main claims and/or conclusions of the paper (regardless of whether the code and data are provided or not)?
    \item[] Answer: \answerYes{} % Replace by \answerYes{}, \answerNo{}, or \answerNA{}.
    \item[] Justification: We provide the implementation details in Appendix. \ref{appendix:exp_setup}.

\item {\bf Open access to data and code}
    \item[] Question: Does the paper provide open access to the data and code, with sufficient instructions to faithfully reproduce the main experimental results, as described in supplemental material?
    \item[] Answer: \answerYes{} % Replace by \answerYes{}, \answerNo{}, or \answerNA{}.
    \item[] Justification: Code will be released after the paper is accepted.

\item {\bf Experimental setting/details}
    \item[] Question: Does the paper specify all the training and test details (e.g., data splits, hyperparameters, how they were chosen, type of optimizer, etc.) necessary to understand the results?
    \item[] Answer: \answerYes{} % Replace by \answerYes{}, \answerNo{}, or \answerNA{}.
    \item[] Justification: We provide the implementation details in Appendix.  \ref{appendix:exp_setup}.

\item {\bf Experiment statistical significance}
    \item[] Question: Does the paper report error bars suitably and correctly defined or other appropriate information about the statistical significance of the experiments?
    \item[] Answer: \answerNo{} % Replace by \answerYes{}, \answerNo{}, or \answerNA{}.
    \item[] Justification: The paper does not report error bars, confidence intervals, or statistical significance information for experiments supporting main claims, due to the high complexity and cost of constructing error bars.

\item {\bf Experiments compute resources}
    \item[] Question: For each experiment, does the paper provide sufficient information on the computer resources (type of compute workers, memory, time of execution) needed to reproduce the experiments?
    \item[] Answer: \answerYes{} % Replace by \answerYes{}, \answerNo{}, or \answerNA{}.
    \item[] Justification: We provide the information for compute resources in Appendix. \ref{appendix:exp_setup}.
    
\item {\bf Code of ethics}
    \item[] Question: Does the research conducted in the paper conform, in every respect, with the NeurIPS Code of Ethics \url{https://neurips.cc/public/EthicsGuidelines}?
    \item[] Answer: \answerYes{} % Replace by \answerYes{}, \answerNo{}, or \answerNA{}.
    \item[] Justification: The research in the paper conforms with the NeurIPS Code of Ethics.

\item {\bf Broader impacts}
    \item[] Question: Does the paper discuss both potential positive societal impacts and negative societal impacts of the work performed?
    \item[] Answer: \answerYes{} % Replace by \answerYes{}, \answerNo{}, or \answerNA{}.
    \item[] Justification: We provide the discussion of broader impacts in Appendix. \ref{appendix:braoder_impacts}.
    
\item {\bf Safeguards}
    \item[] Question: Does the paper describe safeguards that have been put in place for responsible release of data or models that have a high risk for misuse (e.g., pretrained language models, image generators, or scraped datasets)?
    \item[] Answer: \answerNA{} % Replace by \answerYes{}, \answerNo{}, or \answerNA{}.
    \item[] Justification: The benchmark and models utilized in this study do not pose such risks.

\item {\bf Licenses for existing assets}
    \item[] Question: Are the creators or original owners of assets (e.g., code, data, models), used in the paper, properly credited and are the license and terms of use explicitly mentioned and properly respected?
    \item[] Answer: \answerYes{} % Replace by \answerYes{}, \answerNo{}, or \answerNA{}.
    \item[] Justification: All models, code, and data utilized in this study are properly cited to acknowledge their original owners. We have strictly complied with all applicable licenses governing the use of these models, code, and data.

\item {\bf New assets}
    \item[] Question: Are new assets introduced in the paper well documented and is the documentation provided alongside the assets?
    \item[] Answer: \answerNo{} % Replace by \answerYes{}, \answerNo{}, or \answerNA{}.
    \item[] Justification: The paper does not release new assets.

\item {\bf Crowdsourcing and research with human subjects}
    \item[] Question: For crowdsourcing experiments and research with human subjects, does the paper include the full text of instructions given to participants and screenshots, if applicable, as well as details about compensation (if any)? 
    \item[] Answer: \answerNA{} % Replace by \answerYes{}, \answerNo{}, or \answerNA{}.
    \item[] Justification: This paper does not release any new assets.

\item {\bf Institutional review board (IRB) approvals or equivalent for research with human subjects}
    \item[] Question: Does the paper describe potential risks incurred by study participants, whether such risks were disclosed to the subjects, and whether Institutional Review Board (IRB) approvals (or an equivalent approval/review based on the requirements of your country or institution) were obtained?
    \item[] Answer: \answerNA{} % Replace by \answerYes{}, \answerNo{}, or \answerNA{}.
    \item[] Justification: This paper does not involve crowdsourcing activities or research conducted on human subjects.

\item {\bf Declaration of LLM usage}
    \item[] Question: Does the paper describe the usage of LLMs if it is an important, original, or non-standard component of the core methods in this research? Note that if the LLM is used only for writing, editing, or formatting purposes and does not impact the core methodology, scientific rigorousness, or originality of the research, declaration is not required.
    %this research? 
    \item[] Answer: \answerNA{} % Replace by \answerYes{}, \answerNo{}, or \answerNA{}.
    \item[] Justification: The core methods of this paper are unrelated to LLMs and do not involve LLMs as any component.

\end{enumerate}

\end{document}